\documentclass[times, review, 10pt]{elsarticle}

\usepackage{microtype}
\usepackage{graphicx}
\usepackage{subfigure}
\usepackage{booktabs} 
\usepackage{enumitem}
\usepackage{hyperref}

\usepackage{amsmath}
\usepackage{amssymb}
\usepackage{mathtools}
\usepackage{amsthm}

\usepackage{physics}
\usepackage{tikz}
\usetikzlibrary{quantikz2}
\usepackage{booktabs}
\usepackage{multirow} 
\usepackage{natbib}

\journal{Pattern Recognition}

\begin{document}
\begin{frontmatter}

\title{Quantum Simplicial Neural Networks}

\author[xxx]{Simone Piperno}
\author[www]{Claudio Battiloro*}
\author[xxx]{Andrea Ceschini}
\author[www]{Francesca Dominici}
\author[xxx]{Paolo Di Lorenzo}
\author[xxx]{Massimo Panella}

\address[xxx]{Department of Information Engineering, Electronics, and Telecommunications,\\Sapienza University of Rome, Rome, Italy}
\address[www]{Department of Biostatistics, Harvard University, Boston, U.S.A.}

\cortext[cor1]{Corresponding author \{\texttt{cbattiloro@hsph.harvard.edu}\}}

\begin{abstract}
Graph Neural Networks (GNNs) excel at learning from graph-structured data but are limited to modeling pairwise interactions, insufficient for capturing higher-order relationships present in many real-world systems. Topological Deep Learning (TDL) has allowed for systematic modeling of hierarchical higher-order interactions by relying on combinatorial topological spaces such as simplicial complexes. In parallel, Quantum Neural Networks (QNNs) have been introduced to leverage quantum mechanics for enhanced computational and learning power. In this work, we present the first Quantum Topological Deep Learning Model: Quantum Simplicial Networks (QSNs), being QNNs operating on simplicial complexes. QSNs are a stack of Quantum Simplicial Layers, which are inspired by the Ising model to encode higher-order structures into quantum states. Experiments on synthetic classification tasks show that QSNs can outperform classical simplicial TDL models in accuracy and efficiency, demonstrating the potential of combining quantum computing with TDL for processing data on combinatorial topological spaces.
\end{abstract}
\end{frontmatter}

\section{Introduction} \label{sec:intro}
Nowadays, data on graphs are widespread, with applications in social networks, recommendation systems, cybersecurity, sensor networks, and natural language processing. Since their introduction \cite{scarselli2008graph,gori2005new,kipf2016semi,Bruna19,hamilton2017inductive,DuvenaudMABHAA15,GNNGama,gilmer2017neural}, Graph Neural Networks (GNNs) have shown significant results in tasks involving graph data. These networks combine the flexibility of neural networks with the structured knowledge of graph connections. The main approach in GNNs is to derive node feature representations through local aggregation using neighbor information, as determined by the graph's topology. However, GNNs can explicitly model only pairwise interactions, which may not fully represent the multiway interactions in complex systems, like biological networks \cite{lambiotte2019networks}. For this reason, in this paper we will basically rely on the Topological Deep Learning and Quantum Neural Networks in order to extend and generalize the potentialities of GNNs when applied to graph-based tasks. 

Recent advances in Topological Signal Processing (TSP) \cite{barbarossa2020topological, schaub2021signal,sardellitti2022cell,sardellitti2022top,roddenberry2022cellsp, battiloro2023weighted} highlighted the value of modeling data on combinatorial topological spaces, like simplicial or cell complexes, which allow for a more comprehensive representation of higher order interactions. Cell complexes are hierarchical topological spaces composed of a set $\mathcal{V}$ of nodes along with an ensemble of subsets of $\mathcal{V}$, called cells, having some characteristics \cite{grady2010discrete}. Simplicial complexes are particular cell complexes adhering to the so-called inclusion property stating that if a set belongs to the space, then all of its subsets belong to the space. In this case, the involved sets are called simplices, and their order (or dimension) is defined as their cardinality minus one. The advantages shown by TSP techniques have led to the development of neural network architectures for data on such complexes, marking the rise of Topological Deep Learning  (TDL) \cite{hajij2023topological,papamarkou2024positiontdl}.

At the same time, Quantum Neural Networks (QNNs) \cite{tacchinoArtificialNeuronImplemented2019a,chen2020novel,hybrid_ceschini} started to be developed. QNNs offer promising perspectives for surpassing classical neural networks in terms of computational power and efficiency by using the principles of quantum mechanics, such as quantum superposition and entanglement.  For example, \cite{abbasPowerQuantumNeural2021} proved that well-designed QNNs achieve a significantly better effective dimension and can be trained faster than comparable classical neural networks due to their favorable optimization landscapes.  Technologies supporting QNNs range from superconducting machines \cite{tacchino2020quantum} to trapped-ion devices \cite{huber2021realization} and photonic systems \cite{wan2017quantum}. However, despite ongoing improvements in quantum hardware, significant limitations still hinder the full potential of Quantum Machine Learning (QML) \cite{electronics12112379}. Recognizing that fault-tolerant quantum computers are still years away, researchers have made significant progress in developing quantum algorithms tailored for execution on Noisy Intermediate-Scale Quantum (NISQ) devices, that is, quantum computers composed of hundreds of physical qubits without robust error correction protocols \cite{Preskill2018quantumcomputingin}.
Variational Quantum Circuits (VQCs) are currently the most straightforward yet effective way to implement QNNs and take advantage of NISQ hardware \cite{mitarai2018quantum,tacchino2020variational,cerezo2021variational}.
The core concept of VQCs lies in a hybrid quantum-classical approach: the output of the quantum circuit depends on tunable parameterized gates, which are optimized iteratively by a classical processor to minimize a specific loss. 

Within this context, Quantum Graph Neural Networks (QGNNs) represent one of the most promising approaches leveraging the power of quantum computing to process and analyze graph-structured data \cite{qgnn} based on the VQC paradigm. A fundamental aspect of QGNNs lies in the representation of graphs within a quantum computational framework. This involves encoding graph structures into quantum states using quantum circuits and quantum registers, allowing for the manipulation and analysis of graphs with the potential advantages of quantum computing, such as efficient and physics-based Hamiltonian evolution \cite{farhi2014quantum}.
However, the integration of higher-order interactions within QGNNs remains unexplored. Most current research has focused on pairwise interactions and simple graph structures, leaving the study of more complex topological spaces, such as simplicial complexes,  unexplored. 

Our main and original contribution in the paper is the introduction of the first {\bf Quantum TDL (QTDL)} model: {\bf Quantum Simplicial Networks (QSNs)}, which are quantum neural networks operating on simplicial complexes.
A QSN is composed of a stack of Quantum Simplicial Layers (QSLs), with a Multi-Layer Perceptron (MLP) readout. QSLs are designed to encode the underlying structure of the input simplicial complex and the data defined on top of it into quantum states. In particular, we introduce two variants of QSLs, the Base Quantum Simplicial Layer (BQSL) and the Schematic Quantum Simplicial Layer (SQSL). A BQSL is inspired and pseudo-generalizes the Ising Hamiltonian as employed in QGNNs \cite{qgnn}, while a SQSL is a BQSL in which rotation axes are changed to reflect the interaction of simplices of different order.

We explore the theoretical foundations, design, and implementation of the QSN, demonstrating its potential through extensive numerical results. 
In particular, we compared the performance of QSNs against two classical simplicial neural architectures on two synthetic classification tasks. The results demonstrate that QSNs outperform the classical models in both accuracy and parameter efficiency, suggesting that QSNs can substantially enhance the performance of their classical counterparts. This improvement not only paves the way for applying QSNs to real-world scenarios but also opens new avenues for leveraging quantum computing in the analysis and processing of data defined over combinatorial topological spaces. 

The rest of the paper is organized as follows. The most relevant works on simplicial complexes and QGNNs from the literature are outlined in Sect.~\ref{sec:related_works}. The required background on simplicial complexes and quantum computing is introduced in Sect.~\ref{sec:background}. QSNs are thoroughly described and analyzed in Sect.~\ref{sec:quantum_simplicial_layer}. Experimental results are reported in Sect.~\ref{sec:results}. Finally, conclusions and future perspectives on QSNs are delineated in Sect.~\ref{sec:conclusions}.

\section{Related Works}
\label{sec:related_works}

\subsection{Topological Deep Learning on Simplicial Complexes} 
Recent models over simplicial complexes include simplicial convolutional neural networks \cite{ebli2020simplicial}, which generalize graph convolutional neural networks  \cite{kipf2016semi} to SCs. Message-passing neural networks  \cite{gilmer2017neural} were adapted for SCs in \cite{bodnar2021weisfeiler}, in which a test for differentiating isomorphic SCs has also been introduced. This model expands on previous works \cite{bunch2020simplicial, roddenberry2021principled} employing specific simplicial filters \cite{yang2021finite}. \cite{roddenberry2019hodgenet} used recurrent MPNNs for tasks like flow interpolation and graph classification. Further, \cite{yang2021simplicial,yang2023convolutional} developed convolutional architectures able to process interactions across simplicial orders \cite{yang2023convolutional}. \cite{eijkelboom2023mathrmen} explored $E(n)$ equivariant messaging passing in simplicial networks, enabling the usage of geometric features. Simplicial neural networks without message-passing have been introduced in \citep{madhu2024simplicialunsupervised,maggs2024simplicial}. Message-passing neural networks for directed simplicial complexes were presented in \cite{lecha2024dirsnn}.  Attention mechanisms for simplicial neural networks were introduced in \cite{giusti2022simplicial, battiloro2023generalized, anonymous2022SAT, lee2022sgat}. Simplicial Gaussian processes were introduced in \cite{alain2023gaussian, yang2023hodgegaussian}. Finally, \cite{leditto2023qtsp} presented a quantum algorithm for signal filtering on simplicial complexes.

\subsection{Variational Quantum Circuits}
VQCs are low-depth quantum algorithms optimized using a hybrid quantum-classical approach, making them suitable for current small-scale quantum devices \cite{mitarai2018quantum,tacchino2020variational}. The quantum circuit's parametrized gates are iteratively optimized with a classical co-processor, enabling efficient design of low-depth QNNs for tasks like regression and classification \cite{cerezo2021variational}. VQCs make use of a layered encoding and processing in high-dimensional Hilbert space, enhancing expressiveness through hard-to-simulate data encoding \cite{havlicekSupervisedLearningQuantumenhanced2019,schuld2021effect}.
VQCs have shown success in various DL tasks \cite{dallaire2018quantum,9661011} as well as real-world applications \cite{mangini2022quantum,hybrid_ceschini}, showing higher expressive power than classical Deep Neural Networks (DNNs) \cite{zhaoQDNNDeepNeural2021}.

\subsection{Quantum Graph Neural Networks}
QGNNs extend QNNs to graph-structured data, employing quantum registers to represent graph topology, often encoded in the circuit's Hamiltonian \cite{qgnn}. A key aspect of QGNNs lies in graph representation with quantum registers, including techniques for encoding graph structures into quantum states and addressing challenges in manipulating large-scale graphs with limited qubit resources. The underlying idea is to leverage the quantum dynamics to embed data and thus introduce a richer feature map, with characteristics that are hard to access for classical methods \cite{innan2024financial}. The graph topology is encoded in the circuit through the Hamiltonian of the system, where the number of qubits employed is linear with respect to the number of nodes in the graph; this result implies that we can think about quantum circuits with graph-theoretic properties \cite{farhi2014quantum,qgnn}. 
QGNNs' architecture may follow a hybrid quantum-classical scheme \cite{chen2021hybrid,tuysuz2021hybrid}, although relying on classical layers can dilute the benefits of a fully quantum approach. 
QGNNs have shown promise in learning quantum dynamics \cite{qgnn}, particle classification \cite{chen2021hybrid}, and time series prediction \cite{mauro2024hybrid}, though higher-order interactions remain completely unexplored.

\section{Theoretical Background}
\label{sec:background}

\subsection{Remark on Presentation and Notations}\label{sec:notations}
 Originally, the theory of simplicial complexes has been developed within the algebraic topology realm, usually resulting in a notation based on (linear) operators \cite{Hansen2019towardspecsheaf}.  However, the TSP and TDL communities, in agreement with their history, usually employ the matrix representation of the involved linear operators \cite{barbarossa2020topological}. Finally, the QML community usually employs the Dirac notation, being a partial override of the linear operator notation \cite{Nielsen_Chuang_2010}. For this reason, here we decided to keep the matrix notation when presenting TSP/TDL arguments (e.g., Sect.~\ref{sec:backgroundsc}), and the Dirac notation when presenting QTDL arguments (e.g. Sect.~\ref{sec:quantum_simplicial_layer}). This approach, along with a careful choice of the employed jargon, makes the work accessible to a broader and more diverse audience. The above references are useful to clarify the details, if needed. 
\begin{figure}[!ht]
\centering
\includegraphics[width=\textwidth - 1cm]{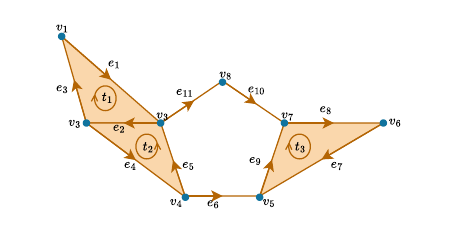} 
\caption{Graphical representation of a order 2 simplicial complex $\mathcal{X}_2$.}
\label{}
\end{figure}

\subsection{Simplicial Complexes}\label{sec:backgroundsc}
Given a finite set of vertices $\mathcal{V}$, a $k$-simplex $\mathcal{H}_{k}$ is a subset of $\mathcal{V}$ with cardinality $k+1$. A face of $\mathcal{H}_{k}$ is a subset with cardinality $k$ and thus a $k$-simplex has $k+1$ faces. A coface of $\mathcal{H}_{k}$ is a $(k + 1)$-simplex
that includes $\mathcal{H}_{k}$ \cite{barbarossa2020topological, lim2020hodge}. 
A simplicial complex $\mathcal{X}_{K}$ of order $K$, is a collection of $k$-simplices $\mathcal{H}_{k}$, $k = 0, \ldots, K$ such that, for any $\mathcal{H}_{k} \in \mathcal{X}_{K}$, $\mathcal{H}_{k-1} \in \mathcal{X}_{K}$ if $\mathcal{H}_{k-1} \subset \mathcal{H}_{k}$ (inclusivity property). We denote the  set of $k$-simplex in $\mathcal{X}_{K}$ as  ${\cal D}_{k} := \{\mathcal{H}_{k}: \mathcal{H}_{k} \in \mathcal{X}_{K}\} $, with $|{\cal D}_{k}| = N_k$ and, obviously, ${\cal D}_{k} \subset {\cal X}_{K}$. We are interested in processing signals defined over a simplicial complex. A $k$-simplicial signal is defined as a mapping from the set of all $k$-simplices contained in the complex to real numbers:
\begin{equation} \label{gen_sig}
s_k: {\cal D}_{k} \rightarrow \mathbb{R}, \,\,\quad k=0, 1, \ldots K.
\end{equation}

The order of the signal is one less the cardinality of the elements of ${\cal D}_{k}$.  A simplicial complex (SC) signal is defined as the concatenation of the signals of each order. In this work, we focus on complexes $\mathcal{X}_{2}$ of order up to two without loss of generality, thus a set of vertices $\mathcal{V}$ with $|\mathcal{V}| = N$, a set of edges $\mathcal{E}$ with $|\mathcal{E}|=E$ and a set of triangles $\mathcal{T}$ with $|\mathcal{T}| = T$ are considered, resulting in ${\cal D}_{0}={\cal V}$ (simplices of order 0), ${\cal D}_{1}={\cal E}$ (simplices of order 1) and ${\cal D}_{2}={\cal T}$ (simplices of order 2). The $k$-simplicial signals are then the following mappings: 
\begin{equation}
s_{0}: {\cal V} \rightarrow \mathbb{R} , \qquad s_{1}: {\cal E} \rightarrow \mathbb{R} , \qquad s_{2}: {\cal T} \rightarrow \mathbb{R} ,
\end{equation}
representing graph, edge, and triangle signals, respectively. We will use the vector notation $\mathbf{s}_0 \in \mathbb{R}^N$, $\mathbf{s}_1 \in \mathbb{R}^E$, $\mathbf{s}_2 \in \mathbb{R}^T$ to denote the collection of the signal values on each 0 (node), 1 (edge), and 2 (triangle) simplex, respectively.  In this case, the corresponding SC signal is given by:
\begin{equation}\label{sc_signal_2}
    \mathbf{s}_{\mathcal{X}} = \big[\mathbf{s}_0, \mathbf{s}_1, \mathbf{s}_2\big] \in \mathbb{R}^{N+E+T}.
\end{equation}

\subsubsection{Algebraic representations} 
The structure of a simplicial complex of order two ${\cal X}_{2}$  is fully described by the set of its incidence matrices $\mathbf{B}_{k}$, $k=1, 2$, given a reference orientation. The entries of the incidence matrix $\mathbf{B}_{k}$ establish which $k$-simplices are incident to which $(k-1)$-simplices.  Denoting the fact that two simplices have the same orientation with $\mathcal{H}_{k-1,i} \sim \mathcal{H}_{k,j}$ and viceversa with $\mathcal{H}_{k-1,i} \not\sim \mathcal{H}_{k,j},$ the entries of $\mathbf{B}_{k}$ are defined as follows:
  \begin{equation} \label{inc_coeff}
  \big[\mathbf{B}_{k} \big]_{i,j}=\left\{\begin{array}{rll}
  0, & \text{if} \; \mathcal{H}_{k-1,i} \not\subset \mathcal{H}_{k,j} \\
  1,& \text{if} \; \mathcal{H}_{k-1,i} \subset \mathcal{H}_{k,j} \;  \text{and} \; \mathcal{H}_{k-1,i} \sim \mathcal{H}_{k,j}\\
  -1,& \text{if} \; \mathcal{H}_{k-1,i} \subset \mathcal{H}_{k,j} \;  \text{and} \; \mathcal{H}_{k-1,i} \not\sim \mathcal{H}_{k,j}\\
  \end{array}\right. .
  \end{equation}
Therefore, $\mathbf{B}_{1} \in \mathbb{R}^{V \times E}$ and  $\mathbf{B}_{2} \in \mathbb{R}^{E \times T}$ model the incidences of nodes to edges and edges to triangles, respectively. From the incidence information, we can build the Hodge Laplacian matrices \cite{goldberg2002combinatorial}, of order $k=0, \ldots, 2$, as follows:
\begin{align}
&\mathbf{L}_{0}=\mathbf{B}_{1}\mathbf{B}_{1}^T, \quad \mathbf{L}_{1}=\underbrace{\mathbf{B}_1^{T}\mathbf{B}_{1}}_{\mathbf{L}_1^{(d)}}+\underbrace{\mathbf{B}_{2}\mathbf{B}_{2}^T}_{\mathbf{L}_1^{(u)}}, \quad \mathbf{L}_{K}=\mathbf{B}_{K}^T\mathbf{B}_{K}.\label{LaplacianK}
\end{align}

The Laplacian of order one contains two terms: the first term $\mathbf{L}^{(d)}_k$, also known as  lower Laplacian, encodes the lower adjacency of $k$-order simplices; the second term $\mathbf{L}_1^{(u)}$, also known as upper Laplacian, encodes the upper adjacency of edges. In particular, two edges are lower adjacent if they share a common vertex, whereas they are upper adjacent if they are faces of a common triangle. Note that the vertices of a graph can only be upper adjacent, if they are incident to the same edge. This is why the graph Laplacian $\mathbf{L}_0$ contains only one term, and it corresponds to the usual graph Laplacian. Hodge Laplacians admit a Hodge decomposition \cite{lim2020hodge},
leading to three orthogonal subspaces. In particular, the $k$-simplicial signal space can be decomposed as:
\begin{equation} \label{hodge_spaces}
\mathbb{R}^{N_{k}} = \text{im}(\mathbf{B}_{k}^T\big) \oplus \text{im}\big(\mathbf{B}_{k+1}\big) \oplus \text{ker}\big(\mathbf{L}_{k}\big)
\end{equation},
where $\oplus$ represents the direct sum of vector spaces.

Thus, any signal $\mathbf{s}_{k}$ of order $k$ can be orthogonally decomposed as:
\begin{equation}
\label{hodge_decomp}
\mathbf{s}_{k}=\underbrace{\mathbf{B}_{k}^T\, \mathbf{s}_{k-1}}_{(a)}+\underbrace{\mathbf{B}_{k+1}\, \mathbf{s}_{k+1}}_{(b)}+\underbrace{\widetilde{\mathbf{s}}_{k}}_{(c)}.
\end{equation}
When edge signals $\mathbf{s}_{1}$ are considered, the Hodge decomposition has a peculiar interpretation that can be found in \cite{barbarossa2020topological, yang2021simplicial}. We usually refer to (a)-(b)-(c) as rotational, solenoidal, and harmonic components of the signal, respectively.

\subsubsection{Simplicial Neural Networks} 
Generalized simplicial convolutional neural networks \cite{yang2023convolutional,battiloro2023generalized} (GSCNs) are a stack of layers, each comprising two main stages: i) a bank of simplicial complex filters \cite{yang2021finite,battiloro2023generalized}, being polynomials of the Laplacians in \eqref{LaplacianK}, and i,i) a point-wise non-linearity. Let us assume that $F_l$ simplicial complex signals 
$\mathbf{Z}_{l} = [\mathbf{Z}_{0,l}, \mathbf{Z}_{1,l}, \mathbf{Z}_{2,l}] \in \mathbb{R}^{(N+E+T)\times F_{l}}$ are given as input to the $l$-th layer of the GSCCN, with $\mathbf{Z}_{0,l}=\{\mathbf{z}_{0,l,f}\}_{f=1}^{F_{l}}\in \mathbb{R}^{N\times F_{l}}$, $\mathbf{Z}_{1,l}=\{\mathbf{z}_{1,l,f}\}_{f=1}^{F_{l}}\in \mathbb{R}^{E\times F_{l}}$, and  $\mathbf{Z}_{2,l}=\{\mathbf{z}_{2,l,f}\}_{f=1}^{F_{l}}\in \mathbb{R}^{T\times F_{l}}$ denoting the signals of each order.  First, each of the input signals is passed through a bank of $F_{l+1}$ polynomial filters. Then, the intermediate outputs  $\{\Tilde{\mathbf{z}}_{k,l,f}\}_f$ are summed to avoid exponential filter growth and, finally, a pointwise non-linearity $\beta_l(\cdot)$ is applied. The output  signals $\mathbf{Z}_{\mathcal{X},l+1}$ of the $l$-th layer read as: 
\begin{align}
\nonumber& \mathbf{Z}_{0,l+1} = \beta_l \Bigg( \overset{J}{\underset{p = 1}{\sum}}(\mathbf{L}_0)^p\mathbf{Z}_{0,l}\mathbf{W}^{(d)}_{l,2p}+ \overset{J }{\underset{p = 0}{\sum}}(\mathbf{L}_0)^p\mathbf{B}_1\mathbf{Z}_{1,l}\mathbf{W}^{(d)}_{l,2p + 1} + \widehat{\mathbf{Q}}_{0}\mathbf{Z}_0\mathbf{W}^{(h)}_l \Bigg),
\end{align}
\begin{align}\label{GSCCN_layer}
 \mathbf{Z}_{1,l+1} = & \beta_l \Bigg( \overset{J}{\underset{p = 1}{\sum}}(\mathbf{L}^{(d)}_1)^p\mathbf{Z}_{1,l}\mathbf{W}^{(d)}_{l,2p} +  \overset{J }{\underset{p = 0}{\sum}}(\mathbf{L}^{(d)}_1)^p\mathbf{B}_1^T\mathbf{Z}_{0,l}\mathbf{W}^{(d)}_{l,2p + 1} \nonumber \\
& +\sum_{p = 1}^{J}(\mathbf{L}^{(u)}_1)^p\mathbf{Z}_{1,l}\mathbf{W}^{(u)}_{l,2p} +  \overset{J }{\underset{p = 0}{\sum}}(\mathbf{L}^{(u)}_1)^p\mathbf{B}_2\mathbf{Z}_{2,l}\mathbf{W}^{(u)}_{l,2p + 1}   
\nonumber +\widehat{\mathbf{Q}}_{1}\mathbf{Z}_1\mathbf{W}^{(h)}_l \Bigg), \nonumber
\end{align}
\begin{align}
 & \mathbf{Z}_{2,l+1} = \beta_l \Bigg( \overset{J}{\underset{p = 1}{\sum}}(\mathbf{L}_2)^p\mathbf{Z}_{2,l}\mathbf{W}^{(u)}_{l,2p}+  \overset{J }{\underset{p = 0}{\sum}}(\mathbf{L}_2)^p\mathbf{B}_2^T\mathbf{Z}_{1,l}\mathbf{W}^{(u)}_{l,2p + 1} +\widehat{\mathbf{Q}}_{2}\mathbf{Z}_2\mathbf{W}^{(h)}_l \Bigg). 
\end{align}

The filters weights $\big\{\mathbf{W}^{(d)}_{l,p}\big\}_{p=1}^{J}$, $\big\{\mathbf{W}^{(u)}_{l,p}\big\}_{p=1}^{J}$ and $\mathbf{W}^{(h)}_l$ are learnable parameters; the order $J$ of the filters, the number $F_{l+1}$ of output signals, and the non-linearity $\beta_l(\cdot)$ are hyperparameters to be chosen (possibly) at each layer. Therefore, a GSCCN of depth $L$ with input data $\mathbf{S} \in \mathbb{R}^{(N+E+T) \times F_0}$ is built as the stack of $L$ layers defined as in \eqref{GSCCN_layer}, where $\mathbf{Z}_{\mathcal{X},0} = \mathbf{S}$.  Based on the learning task at hand, an additional read-out layer can be inserted after the last GSCCN layer. Intuitively, in a GSSCN layer, simplices of the same order interact with each other based on the upper and lower neighborhoods induced by the Hodge Laplacians, while simplices of different order interact based on the neighborhoods induced by the incidence relations. Please refer to \cite{battiloro2023generalized} for further details.

\subsection{Quantum Neural Networks}\label{sec:backgroundqnn}
\begin{figure*}[!ht]
    \centering
    \includegraphics[width=1\linewidth-2cm]{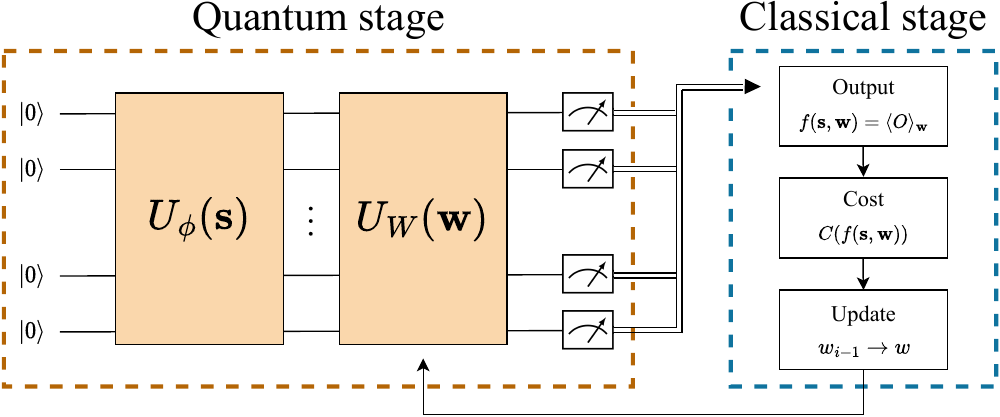}
    \caption{Functioning scheme of a VQC.}
    \label{fig:VQC_diagram}
\end{figure*}

A QNN  processes vector data $\mathbf{s} \in \mathbb{R}^N$ using VQCs, as shown in Fig.~\ref{fig:VQC_diagram}.

The QNN framework consists of three key components: data encoding, the application of an expressive ansatz to the quantum state, and measurement operations. A quantum feature map $\phi: \mathbb{R}^N \rightarrow H^{2^N}$ is employed by applying a unitary operator $U_{\phi}(\mathbf{s})$ to the initial $\ket{0}^{\otimes N}$ state to encode classical data $\mathbf{s} \in \mathbb{R}^N$ into an $n$-qubit quantum circuit:
\begin{equation}		
	U_{\phi}(\mathbf{s})\ket{0}^{\otimes N} = \ket{\phi(\mathbf{s})} = \ket{\psi}\,, 
\end{equation}
where the notation $\ket{0}^{\otimes N}$ represents a tensor product of N identical copies of the initial quantum state $\ket{0}$:
\begin{equation}
    \ket{0}^{\otimes N} =  \ket{0}_1 \otimes \ket{0}_2 \otimes ... \otimes \ket{0}_N.
\end{equation}

A practical quantum advantage cannot be achieved without a robust data encoding strategy that effectively maps classical features into quantum states \cite{weigold2021expanding}. For this reason, the selection of the unitary operator $U_{\phi}(\cdot)$ is critical, as it directly affects the performance of the underlying QNN. 

Among the numerous encoding methods available \cite{schuld2021machine}, angle encoding, which uses parameterized rotation gates, is particularly favored for its effectiveness in encoding continuous variables into quantum states \cite{havlicekSupervisedLearningQuantumenhanced2019, zhaoQDNNDeepNeural2021}. It maps each input feature onto a qubit via quantum rotation gates, enabling the representation of $n$ input features using $n$ qubits. 
After the data encoding phase, an ansatz $U_W(\mathbf{w})$, composed of {$\mathbf{w}$-parametrized} unitaries, with $\mathbf{w} \in \mathbb{R}^F$, is applied to the quantum state $\ket{\psi}$:
\begin{equation}	
\label{eq:ansatz}
	U_W(\mathbf{w})\ket{\psi} = \ket{\psi'},
\end{equation}
with $\ket{\psi'}$ being the evolved state.
The unitaries in $U_W{(\mathbf{w})}$ consist of tunable single-qubit rotation gates and two-qubit entangling gates, which are randomly initialized. 

Rotation gates are crucial for navigating the Hilbert space, while the entangling gates generate strong correlations among the qubits, creating entangled quantum registers. The parameters of these gates are optimized using a classical co-processor to minimize a  loss function. To enhance the expressiveness of the models, layers of rotations and entanglements may be iteratively applied within the ansatz \cite{abbasPowerQuantumNeural2021,cerezoansatz}, resulting in:
\begin{equation}	
\label{eq:data re-uploading}
	U_W(\mathbf{w}) = \prod_{l=1}^L{U_l(\mathbf{w}^{l})},
\end{equation}
where $\mathbf{w}^{l} \in \mathbb{R}^{F_l}$ refers to the parameters in the $l$-th layer of the ansatz.

Multiple circuit shots are necessary to obtain a reliable output due to the inherent probabilistic nature of quantum mechanics. For this reason, we measure the expectation value $\langle \hat{O} \rangle = \bra{\psi'}\hat{O}\ket{\psi'}$ of the desired operator $\hat{O}$ on the final state, thus getting:
\begin{equation}	
 f(\mathbf{s}, \mathbf{w}) = \bra{\phi(\mathbf{s})}U_W(\mathbf{w})^{\dagger} \, \hat{O} \,\, U_W(\mathbf{w})\ket{\phi(\mathbf{s})}\,.
 \end{equation}
In Dirac notation, $\dagger$ represents the Hermitian adjoint (or conjugate transpose) of the operator $U_W$:
\begin{equation}
    U_W^\dagger=\overline{U_W}^T
\end{equation}
where $\overline{U_W}$ denotes the complex conjugate and $T$ the transpose. The notation $\bra{\phi}$ indicates a bra, which is the dual vector associated with the quantum state $\ket{\phi}$, hence we have $\bra{\phi} = (\ket{\phi})^\dagger$.
Lastly, the product $\bra{\phi}A\ket{\phi}$ represents the expectation value of the operator A in the state $\ket{\phi}$.

Here, $\hat{O}$ can be any hermitian operator acting on the $n$-qubit system. In practice, $\hat{O}$ is often a tensor product of Pauli operators.
By measuring the qubits in the Pauli-Z basis, we can determine the expectation value with respect to the Pauli-Z operator $\sigma_z = \begin{psmallmatrix}
1 & 0\\
0 & -1
\end{psmallmatrix}$.
This way, the output range of the QNN for a single qubit is constrained to $[-1, 1]$, reflecting the possible measurement outcomes for a single qubit in the Pauli-Z basis, where $-1$ and $1$ are the eigenvalues associated with the $\ket{1}$ and $\ket{0}$ states, respectively. 

The outcome $f(\mathbf{s}, \mathbf{w})$ of the QNN is then passed to a suitable loss function; classical optimization routines are used to update $\mathbf{w}$ and train the QNN. To compute gradients with respect to quantum parameters $\mathbf{w}$, the parameter-shift rule for the $i$-th parameter $w_i$ is commonly employed:
\begin{equation}
\nabla_{w_i}f(\mathbf{s},w_i) = \frac{1}{2} \left[ f\bigl(\mathbf{s},w_i + \frac{\pi}{2}\bigr) - f(\mathbf{s},w_i - \frac{\pi}{2}\bigr) \right],
\end{equation}
where $f(\mathbf{x}, w_i)$ is the quantum circuit output with $w_i$ being the parameter of interest, while the other parameters are kept fixed.

\section{Quantum Simplicial Neural Network}
\label{sec:quantum_simplicial_layer}
In this section, we introduce QSNs, i.e. the first QNN operating on simplicial complexes, being a stack of QSLs. We first describe the general structure of a QSL, and then we particularize it to the Base QSL and the Schematic QSL. We present QSNs for simplicial complexes of order up to two for the sake of exposition. 
However, due to the way they are defined, only the QSNs employing the Base QSLs can be straightforwardly generalized to complexes of arbitrary order, while the Schematic QSLs do not lend themselves to such direct generalization.
\subsection{QSN Architecture}
\label{layer_arch}
\begin{figure}[!ht]
\centering
\includegraphics[width=\textwidth - 1cm]{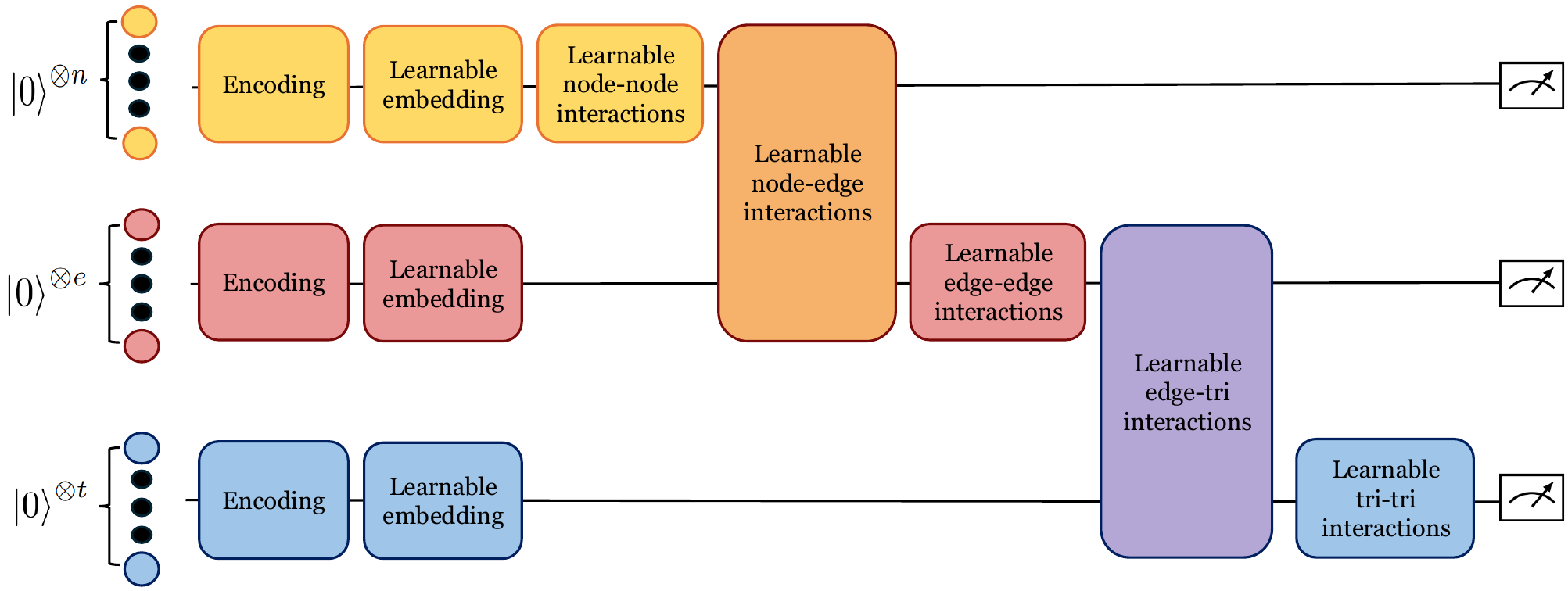} 
\caption{The Quantum Simplicial Layer scheme.}
\label{schema}
\end{figure}

A QSN is a stack of QSLs plus an MLP readout layer. The scheme of a QSL is depicted in Fig.~\ref{schema}.

The layer is designed with the goal of mapping all the interactions among simplices in a simplicial complex into a quantum circuit.  In particular, we consider a simplicial complex of order two $\mathcal{X}_2 $ composed of  $N$ vertices, $E$ edges, and $T$ triangles and a single input simplicial complex signal $\mathbf{s}$ defined as in \eqref{sc_signal_2}.  In the following, we denote the global state of the circuit as $\ket{\Psi}$. The main components of a QSL are:
\begin{enumerate}[leftmargin=*]
    \item \textbf{Qubits Initialization.} The $N+E+T$ qubits are initialized in the $\ket{0}$ state. Formally, the initial global state of the circuit can be written as the tensor product of single-qubit rotations:
    \begin{equation}
    \label{init}
        \ket{\Psi}^{\text{init}} = \ket{0}^{\otimes (N+E+T)}.
    \end{equation}
    \item \textbf{Encoding.} Then, we encode in the circuit the signal we have on each simplex such that $\ket{\Psi}^{\text{enc}} = U_{\phi}(\mathbf{s})\ket{\Psi}^{\text{init}}$, thus each qubit is rotated along a given axis by an angle between $[-\pi, \pi]$ proportional to the input feature of the simplex. Formally, the encoding global state of the circuit can be written as:
    \begin{align}\label{enc}
    \begin{split}
        \ket{\Psi}^{\text{enc}} = \left( \bigotimes_{i=1}^N R_{k_1}(\pi s_i) \right) & \otimes \left( \bigotimes_{i=N+1}^{N+E} R_{k_2}(\pi s_i) \right) \otimes \left( \bigotimes_{i=N+E+1}^{N+E+T} R_{k_3}(\pi s_i) \right)  \ket{\Psi}^{\text{init}},
    \end{split}
    \end{align} 
    where $s_i$ is the $i$-th component of $\mathbf{s}$, i.e. the signal of the $i$-th simplex, $R_{k_1}(\pi s_i)$, $R_{k_2}(\pi s_i)$, and $R_{k_3}(\pi s_i)$ are the common rotation gates applied to qubits corresponding to nodes, edges, and triangles, respectively, and $k_1, k_2, k_3 \in \{x,y,z\}$ depending on the version of the layer employed. Lastly, $\otimes$ represents the tensor product operation.
    \item \textbf{Learnable Embedding.} The updated embeddings of the simplices, i.e., the updated global state of the circuit, are obtained by rotating each qubit along a given axis - which could be different from the one of the encoding - by a learnable angle, such that $\ket{\Psi}^{\text{emb}} = U_{W_1}(\mathbf{W})\ket{\Psi}^{\text{enc}}$. Formally, the embedding global state of the circuit can be written as:
    \begin{align}\label{emb}
        \begin{split}
            \ket{\Psi}^{\text{emb}} = \left( \bigotimes_{i=1}^N R_{k_4}(\mathbf{W}_{i,i}) \right) & \otimes 
        \left( \bigotimes_{i=N+1}^{N+E} R_{k_5}(\mathbf{W}_{i,i}) \right)\otimes \left( \bigotimes_{i=N+E+1}^{N+E+T} R_{k_6}(\mathbf{W}_{i,i}) \right) \ket{\Psi}^{\text{enc}}
        \end{split},
    \end{align}
    where $k_4, k_5, k_6 \in \{x,y,z\}$ depending on the version of the layer employed, and $\mathbf{W}$ is a matrix of learnable parameters having shape $(N+E+T, \,N+E+T)$, which allows to model simplexes interactions.
    \item \textbf{Learnable Interactions.} We leverage the interactions among simplices induced by the neighborhoods of the simplicial complex in a learnable fashion by using LI$_{k,p}$ gates, with $k,p \in \{x,y,z\}$, defined as:
    \begin{align}\label{LI}
    \begin{split}
         \text{LI}_{k,p} (\mathbf{W}_{i,j}) = 
        \text{CX} & \cdot (\mathbb{I} \otimes R_{k,p} (\mathbf{W}_{i,j})) \cdot \text{CX}
    \end{split}
    \end{align}
    and which is represented in Fig.~\ref{fig:circuits}a, such that $\ket{\Psi}^{\text{LI}} = U_{W_2}(\mathbf{W})\ket{\Psi}^{\text{emb}}$. Suppose $ \text{LI}_{k,p}$ is applied to qubits $i$ and $j$: here, CX represents the controlled-NOT quantum operation, while $\mathbf{W}_{i,j}$ is the parameter related to the simplex involving qubits $i$ and $j$.
    LI$_{k,p}$ gate models the two way interaction between simplex (qubit) $i$ and simplex (qubit) $j$, and $R_{k,p}$ is defined such that
    \begin{equation}
    R_{k,p} (\mathbf{W}_{i,j})=
    \begin{cases}
    R_{k} (\mathbf{W}_{i,j}) \text{ if }k=p,\\
    R_{p} (\mathbf{W}_{i,j}) R_{k} (\mathbf{W}_{i,j}) \text{ oth.}\\
    \end{cases}
    \end{equation}

    Therefore, $R_{k,p}$ is simply a rotation $R_{k}$  when $k=p$, as in Fig.~\ref{fig:circuits}b,  while it is a concatenation of two rotations of the same angle on the 2 axes when $k\neq p$ as shown in Fig.~\ref{fig:circuits}c. 
    This implementing choice reflects the design principle of the schematic architecture, which aims to "register the same information on the axes involved." This ensures that the same information is effectively transferred across the relevant components.
%
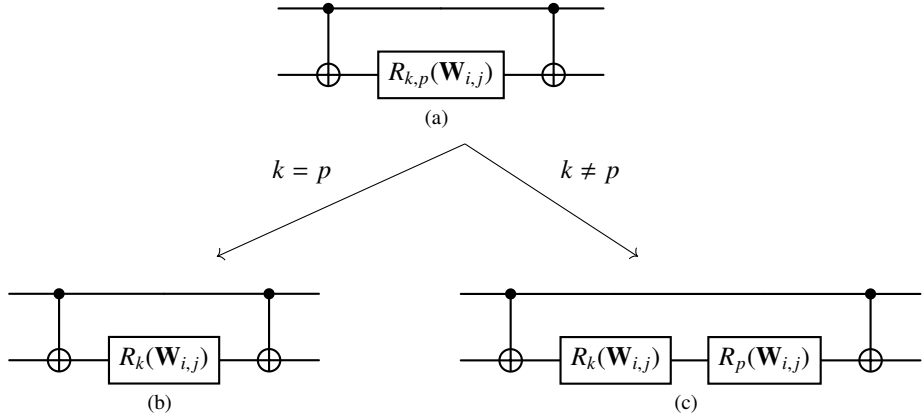
\begin{figure}[!ht]
    \centering
    \subfigure[]{
    \begin{quantikz}
    & \ctrl{1} & & \ctrl{1} & \\
    & \targ{} & \gate{R_{k,p}(\mathbf{W}_{i,j})} &\targ{} &
    \end{quantikz}
    }\\ 
    
    \begin{tikzpicture}
        \node at (-3.5,-0.4) {
            \subfigure[]{
            \begin{quantikz}
            & \ctrl{1} & & \ctrl{1} & \\
            & \targ{} & \gate{R_k(\mathbf{W}_{i,j})} &\targ{} &
            \end{quantikz}
            }
        };
        
        \node at (3.5,-0.4) {
            \subfigure[]{
            \begin{quantikz}
            & \ctrl{1} & & & \ctrl{1} & \\
            & \targ{} & \gate{R_k(\mathbf{W}_{i,j})} & \gate{R_p(\mathbf{W}_{i,j})} &\targ{} &
            \end{quantikz}    
            }
        };
        
        \draw[->] (0.5,2.3) -- (-2.8,0.8) node[midway, above left, yshift=3pt] {$k=p$};
        \draw[->] (0.5,2.3) -- (2.8,0.8) node[midway, above right, yshift=3pt] {$k \neq p$};
    \end{tikzpicture}
    
    \caption{Circuital representation of $\text{LI}_{k,p}$ (a), $\text{LI}_{k,p}$ with $k=p$ (b), and $\text{LI}_{k,p}$ with $k \neq p$ (c).}
    \label{fig:circuits}
\end{figure}

    Mimicking the rationale of classical simplicial neural networks as the GSCNN in \eqref{GSCCN_layer}, in a QSN we enable simplices of the same and different order to interact with each other based on the upper and lower neighborhoods induced by the Hodge Laplacians and the neighborhood induced by the incidence relations, respectively. Formally, the updated global state of the circuit can be written as:

\begin{align}\label{eq:global_LI_state}
\begin{split}
    \ket{\Psi}^{\text{LI}} =& 
    \prod_{\substack{1 \leq i < j \leq T \\ [\mathbf{L}^{(d)}_2]_{i,j} \neq 0}} 
    \Big( \bigotimes_{n=1}^{N+E+T} O_n^{(N+E+i,N+E+j)} \Big) 
    \prod_{\substack{1 \leq i \leq E \\ 1 \leq j \leq T \\ [\mathbf{B}_2]_{i,j} \neq 0}}
    \Big( \bigotimes_{n=1}^{N+E+T} O_n^{(N+i, N+E+j)} \Big)\\
    & \prod_{\substack{1 \leq i < j \leq E \\ [\mathbf{L}^{(u)}_1]_{i,j} \neq 0}} 
    \Big( \bigotimes_{n=1}^{N+E+T} O_n^{(N+i, N+j)} \Big)
    \prod_{\substack{1 \leq i < j \leq E \\ [\mathbf{L}^{(d)}_1]_{i,j} \neq 0}} 
    \Big( \bigotimes_{n=1}^{N+E+T} O_n^{(N+i,N+j)} \Big)\\
    &\prod_{\substack{1 \leq i \leq N \\ 1 \leq j \leq E \\ [\mathbf{B}_1]_{i,j} \neq 0}} 
    \Big( \bigotimes_{n=1}^{N+E+T} O_n^{(i, N+j)} \Big) 
    \prod_{\substack{1 \leq i < j \leq N \\ [\mathbf{L}^{(u)}_0]_{i,j} \neq 0}} \Big( \bigotimes_{n=1}^{N+E+T} O_n^{(i,j)} \Big) 
    \ket{\Psi}^{\text{emb}},
\end{split}
\end{align}
where \(\bigotimes_{n=1}^{N+E+T} O_n^{(i,j)}\) represents the tensor product over all qubits in the quantum register and is a unitary matrix of dimension $2^{N+E+T} \times 2^{N+E+T}$, while \(O_n^{(i,j)}\) is defined as:
  \begin{equation}
    O_n^{(i,j)} = 
      \begin{cases}
      \text{LI}_{k,p}(\mathbf{W}_{i,j}), & \text{if } n \in \{i,j\}, \\
      \mathbb{I}, & \text{otherwise},
      \end{cases}
  \end{equation}
  and the products over the subsets apply the appropriate \(\text{LI}\) interactions. In \eqref{eq:global_LI_state}, the first term applied to $\ket{\Psi}^{\text{emb}}$ leverages node-node interactions (encoded in $\mathbf{L}^{(u)}_0$), the second term leverages node-edge interactions (encoded in $\mathbf{B}_1$), the third and fourth terms leverage lower and upper edge-edge interactions (encoded in $\mathbf{L}^{(d)}_1$ and $\mathbf{L}^{(u)}_1$, respectively), the fifth leverages edge-triangle interactions (encoded in $\mathbf{B}_2$), and the sixth term leverages triangle-triangle interactions (encoded in $\mathbf{L}^{(d)}_2$). The $i < j$ constraint in the first, third, fourth, and sixth terms in \eqref{eq:global_LI_state} avoids using twice the interaction among the same pairs of simplices and removes self-simplex interactions, which are already used in the embedding phase. Each layer of interactions accounts for specific simplex relationships within the simplicial complex, following the rationale of classical simplicial neural networks.

    \item \textbf{Measurement.} 
%
After the application of all learnable interactions, resulting in the final state \(\ket{\Psi}^{\text{LI}}\), we proceed to the measurement stage. In this step, each qubit corresponding to a simplex is individually measured to extract meaningful embeddings that represent the updated state of each simplex. The measurement is performed by calculating the expectation value of the Pauli-Z operator, $\sigma_z$, for each qubit. This expectation value provides a real-valued output that serves as the updated representation for each simplex in the simplicial complex.
Formally, for each qubit $i$, the measurement is defined as:
\begin{align}\label{measurement}
    \hat{y}_i = \langle \Psi^{\text{LI}} | \sigma_z^{(i)} | \Psi^{\text{LI}} \rangle, \quad i = 1, \dots, N + E + T,
\end{align}
where $\ket{\Psi}^{\text{LI}}$ is the final global state of the quantum circuit after applying all learnable interactions and
$\sigma_z^{(i)}$ denotes the Pauli-Z operator acting on the $i$-th qubit.

The measurement of all qubits results in a vector of embeddings that represents the state of all simplices in the complex. This vector is given by:
\begin{align}
    \mathbf{\hat{y}} = \left( \langle \Psi^{\text{LI}} | \sigma_z^{(1)} | \Psi^{\text{LI}} \rangle, \langle \Psi^{\text{LI}} | \sigma_z^{(2)} | \Psi^{\text{LI}} \rangle, \dots, \langle \Psi^{\text{LI}} | \sigma_z^{(N+E+T)} | \Psi^{\text{LI}} \rangle \right),
\end{align}
where each component \(\langle \Psi^{\text{LI}} | \sigma_z^{(i)} | \Psi^{\text{LI}} \rangle\) provides the measured value for the $i$-th qubit, corresponding to the $i$-th simplex.
Such a vector can be further processed according to the requirements of downstream tasks. For example, these embeddings can be fed into additional layers of a hybrid quantum-classical architecture, or serve as input features for classical machine learning models.     
\end{enumerate}

The initialization of qubits and the measurement process occur exclusively at the very first and very last layer of the QSN, respectively. Notably, by integrating the encoding step within each individual layer, we seamlessly enable data reuploading \cite{reuploading} as we stack multiple layers in the network. 
By representing the simplices in this way, we capture the complex interaction patterns encoded in the quantum circuit, which reflect the structure of the original simplicial complex.
In the following sections, we introduce two versions of a QSL.

\subsection{Base Quantum Simplicial Neural Network}
The first layer we introduced is the Base Quantum Simplicial Network (BQSN). This version of the network is inspired from \cite{qgnn} and exploits the gates used for describing the Ising model. 
In particular we take inspiration from the section 4.1 \emph{``Learning Quantum Hamiltonian Dynamics with Quantum Graph Recurrent Neural Networks''}. They describe the target Hamiltonian $\hat{H}$ as:
$$
\hat{H} = \sum_{\{j,k\}\in \mathcal{E}} J_{jk} \hat{Z}_j \hat{Z}_k + \sum_{v\in V} Q_v \hat{Z}_v + \sum_{v\in V} \hat{X}_j
$$
in which $V$ and $\mathcal{E}$ represent respectively the set of nodes and edges, while $\{J_{jk},Q_v\}$, where ${\{j,k,v\}\in V}$, the learnable parameters that optimize the target Hamiltonian. \newline
For our BQSN we adapted this type of encoding, embedding and interaction effect to our scheme, which uses one qubit per simplex, as described in \ref{layer_arch}. These operations are performed in a systematic way and work as follows.  
\begin{itemize}[leftmargin=*]
    \item \textbf{Encoding.} Starting from $\ket{\Psi}^{\text{init}}$, each qubit is assigned to a node, a edge or a triangle; it is rotated along the $x$ axis by an angle between $[-\pi, \pi]$, proportional to the input feature. Then, \eqref{enc} becomes
    \begin{align}
    \begin{split}
            \ket{\Psi}^{\text{enc}} =
            \Big( \bigotimes_{i=1}^{N+E+T} R_{x}(\pi s_i) \Big) \ket{\Psi}^{\text{init}}
    \end{split}
    \end{align}
    
    \item \textbf{Embedding.} The embedding element is obtained by rotating along the $z$-axis by a learnable angle. In this case, \eqref{emb} becomes:
    \begin{align}
    \begin{split}
            \ket{\Psi}^{\text{emb}} = \Big( \bigotimes_{i=1}^{N+E+T} R_{z}(\mathbf{W}_{i,i}) \Big) \ket{\Psi}^{\text{enc}}.
    \end{split}
    \end{align}
    
    \item \textbf{Interactions.} Interactions are implemented by the use of $R_{z,z}$ gates between the interacting simplexes (qubits). Accordingly, \eqref{eq:global_LI_state} develops as:
    %
\begin{align}\label{eq:global_LI_state2}
\begin{split}
    \ket{\Psi}^{\text{LI}} =& 
    \prod_{\substack{1 \leq i < j \leq T \\ [\mathbf{L}^{(d)}_2]_{i,j} \neq 0}} 
    \Big( \bigotimes_{n=1}^{N+E+T} O_n^{(N+E+i,N+E+j)} \Big) 
    \prod_{\substack{1 \leq i \leq E \\ 1 \leq j \leq T \\ [\mathbf{B}_2]_{i,j} \neq 0}}
    \Big( \bigotimes_{n=1}^{N+E+T} O_n^{(N+i, N+E+j)} \Big)\\
    & \prod_{\substack{1 \leq i < j \leq E \\ [\mathbf{L}^{(u)}_1]_{i,j} \neq 0}} 
    \Big( \bigotimes_{n=1}^{N+E+T} O_n^{(N+i, N+j)} \Big)
    \prod_{\substack{1 \leq i < j \leq E \\ [\mathbf{L}^{(d)}_1]_{i,j} \neq 0}} 
    \Big( \bigotimes_{n=1}^{N+E+T} O_n^{(N+i,N+j)} \Big)\\
    &\prod_{\substack{1 \leq i \leq N \\ 1 \leq j \leq E \\ [\mathbf{B}_1]_{i,j} \neq 0}} 
    \Big( \bigotimes_{n=1}^{N+E+T} O_n^{(i, N+j)} \Big) 
    \prod_{\substack{1 \leq i < j \leq N \\ [\mathbf{L}^{(u)}_0]_{i,j} \neq 0}} \Big( \bigotimes_{n=1}^{N+E+T} O_n^{(i,j)} \Big) 
    \ket{\Psi}^{\text{emb}},
\end{split}
\end{align}
where \(\bigotimes_{n=1}^{N+E+T} O_n^{(i,j)}\) represents the tensor product over all qubits in the quantum register, and \(O_n^{(i,j)}\) is defined as:
  \begin{equation}
    O_n^{(i,j)} = 
      \begin{cases}
      \text{LI}_{z,z}(\mathbf{W}_{i,j}), & \text{if } n \in \{i,j\}, \\
      \mathbb{I}, & \text{otherwise},
      \end{cases}
  \end{equation}
\end{itemize}
  
\subsection{Schematic Quantum Simplicial Network}
The next network we introduced is the Schematic Quantum Simplicial Network (SQSN). Here we simply change the rotation axes. The undelying idea at the base of these changes is that by encoding all the information about a simplical order along a given axis allows to describe different order interactions by the interplay of different axes. To achieve this purpose, nodes are associated to the $x$-axis, edges to the $y$-axis and triangles to the $z$-axis, hence the previous elements are modified as follows.
\begin{itemize}
    \item \textbf{Encoding.} 
    Starting from $\ket{\Psi}^{\text{init}}$, each qubit is assigned to a node, a edge or a triangle; it is rotated along the associated axis by an angle between $[-\pi, \pi]$, proportional to the input feature. For the SQSN, \eqref{enc} becomes:
    \begin{align}
    \begin{split}
        \ket{\Psi}^{\text{enc}} = \left( \bigotimes_{i=1}^N R_{x}(\pi s_i) \right) & \otimes \left( \bigotimes_{i=N+1}^{N+E} R_{y}(\pi s_i) \right) \otimes \left( \bigotimes_{i=N+E+1}^{N+E+T} R_{z}(\pi s_i) \right)
        \ket{\Psi}^{\text{init}}.
    \end{split}
    \end{align} 
    \item \textbf{Embedding.} The embedding element is obtained by rotating each qubit $i$ along the same axis it was rotated in the encoding, this time by a learnable angle $\mathbf{W}_{i,i}$.
    We can express \eqref{emb} as:
    \begin{align}
        \begin{split}
            \ket{\Psi}^{\text{emb}} = \left( \bigotimes_{i=1}^N R_{x}(\mathbf{W}_{i,i}) \right) & \otimes 
        \left( \bigotimes_{i=N+1}^{N+E} R_{y}(\mathbf{W}_{i,i}) \right)\otimes \left( \bigotimes_{i=N+E+1}^{N+E+T} R_{z}(\mathbf{W}_{i,i}) \right) \ket{\Psi}^{\text{enc}}
        \end{split}.
    \end{align}

    \item \textbf{Interactions.} Interactions are encoded by the use of the previously introduced $\text{LI}_{k,p}$ gates \ref{LI}.
    Consequently, \eqref{eq:global_LI_state} becomes:
    %
    \begin{align}\label{eq:global_LI_state3}
    \begin{split}
    \ket{\Psi}^{\text{LI}} =& 
    \prod_{\substack{1 \leq i < j \leq T \\ [\mathbf{L}^{(d)}_2]_{i,j} \neq 0}} 
    \Big( \bigotimes_{n=1}^{N+E+T} O_{n,z,z}^{(N+E+i,N+E+j)} \Big) 
    \prod_{\substack{1 \leq i \leq E \\ 1 \leq j \leq T \\ [\mathbf{B}_2]_{i,j} \neq 0}}
    \Big( \bigotimes_{n=1}^{N+E+T} O_{n,y,z}^{(N+i, N+E+j)} \Big)\\
    & \prod_{\substack{1 \leq i < j \leq E \\ [\mathbf{L}^{(u)}_1]_{i,j} \neq 0}} 
    \Big( \bigotimes_{n=1}^{N+E+T} O_{n,y,y}^{(N+i, N+j)} \Big)
    \prod_{\substack{1 \leq i < j \leq E \\ [\mathbf{L}^{(d)}_1]_{i,j} \neq 0}} 
    \Big( \bigotimes_{n=1}^{N+E+T} O_{n,y,y}^{(N+i,N+j)} \Big)\\
    &\prod_{\substack{1 \leq i \leq N \\ 1 \leq j \leq E \\ [\mathbf{B}_1]_{i,j} \neq 0}} 
    \Big( \bigotimes_{n=1}^{N+E+T} O_{n,x,y}^{(i, N+j)} \Big) 
    \prod_{\substack{1 \leq i < j \leq N \\ [\mathbf{L}^{(u)}_0]_{i,j} \neq 0}} \Big( \bigotimes_{n=1}^{N+E+T} O_{n,x,x}^{(i,j)} \Big) 
    \ket{\Psi}^{\text{emb}},
    \end{split}
    \end{align}
    where \(\bigotimes_{n=1}^{N+E+T} O_{n,k,p}^{(i,j)}\) represents the tensor product over all qubits in the quantum register, and \(O_{n,k,p}^{(i,j)}\) is defined as:
      \begin{equation}
        O_{n,k,p}^{(i,j)} = 
          \begin{cases}
          \text{LI}_{k,p}(\mathbf{W}_{i,j}), & \text{if } n \in \{i,j\}, \\
          \mathbb{I}, & \text{otherwise},
          \end{cases}
      \end{equation}
\end{itemize}

Such introduced networks will have the same number of parameters for the same input graph, which makes them easily comparable in the experiments, which are reported in the next sections. We report in Table~\ref{tab: layers_diff} a summary of the differences between the two proposed architectures.

\begin{table}[!ht] 
\caption{Comparison of network architectures.} 
\label{tab: layers_diff}
\vspace{6pt}
\centering
\footnotesize
\begin{tabular}{lcc}
\toprule
Function & BQSN & SQSN \\
\midrule
Encoding & $R_x$ &  $\{R_x, R_y, R_z\}$  \\
Embedding &  $R_z$ & $\{R_x, R_y, R_z\}$ \\
Interactions &  $R_{zz}$ & $R_{i,j}$\\
\bottomrule
\end{tabular}
\end{table}

\subsection{Complexity Analysis}
The number of parameters of the architecture changes accordingly to the graph we are dealing with because we are defining our layers on its structure. If we call $p_n, \, p_e, \, p_t$ the number of parameters needed to encode the information about nodes, edges and triangles respectively we can see that:
\begin{equation}
    p_n = n + C(\mathbf{L_0}) + C (\mathbf{B_1})\,,
\end{equation}
where $C$ is a function that returns the number of elements greater than zero in the matrix it is given in input. We can see in this case we are accounting for the $n$ (one per node) learnable parameters of the embedding, the second term accounts for the node-node interaction and the last term for the node-edge interactions. 

Analogously, we can define:
\begin{equation}
p_e = e + C(\mathbf{L_1^{(d)}}) + C(\mathbf{L_1^{(u)}}) + C(\mathbf{B_2})\,,
\end{equation}
where the first term accounts for the embedding of the edges, the second and third for the edge-edge interactions and the last for the edge-triangle interactions.
For the parameters of the triangles we'll have:
\begin{equation}
p_t = t + C(\mathbf{L_2})\,,
\end{equation}
where the first term is for the triangles' embedding and the last for the triangle-triangle interaction.
The number of parameters $p$ for a single layer will then be:
\begin{equation}
p = p_n+p_e+p_t\,.
\end{equation}


\section{Numerical Results}
\label{sec:results}
We test the effectiveness of QSNs on two synthetic tasks: edge-level solenoidal component detection and source localization.
We compare five different architectures: three classical and two quantum. The three classical architectures are a GSCN as in \eqref{GSCCN_layer}, its attentional version, i.e., a Generalized Simplicial Attention Neural Network (GSAN) \cite{battiloro2023generalized} and a MLP.
The quantum architectures are QSNs with BQS and SQS layers, referred to as BQSN and SQSN, respectively. \textcolor{black}{Additionally, for the source localization task, we also incorporate a sixth architecture, a QGNN, which is considered for comparison purposes.}

In order to ensure fairness in the comparison, we use configurations with a similar number of learnable parameters. Please notice that the number of parameters of the QSNs is the same for the two versions and it is dataset-dependent, in the sense that it depends on the number of simplices in the considered simplicial complex; this is why we cannot perfectly match the number of parameters of quantum and classical SNs, being the number of parameters of the former independent of the number of simplices. Each model is tested varying the number of layers from 1 to 5.

\subsection{Solenoidal Component Detection} 
\label{synthetic_data1}

We consider a synthetic edge signal classification task. We generate a random simplicial complex of order 2 having a minimum number of simplices equal to 12 and a maximum number of simplices equal to 16. In particular, we generate the complex hierarchically: we first generate the nodes by considering an average number of 10, then we generate edges with a probability of 0.3 and finally fill triangles with a probability of 0.8. At this point, we generate a dataset of 1000 signals defined on the complex as follows. Edge signals (i.e. $k=1$ in \eqref{gen_sig}) are generated via a simplified variation of the model in \eqref{hodge_decomp}: each of the edge signals is generated as
\begin{equation}\label{eq:sign_synth}
    \mathbf{s}_{1}=\mathbf{B}_{1}^T\, \mathbf{s}_{0}+A\mathbf{B}_{2}\, \mathbf{s}_{2}\,,
\end{equation}
where $\mathbf{s}_{0}$ and $\mathbf{s}_{2}$  are random vectors whose entries are i.i.d. from the standard normal distribution, and $A$ is Bernoulli random variable, with probability of success ($A=1$) set to 0.5. In other words, we generate edge signals with an irrotational component, no harmonic component, and a randomly assigned solenoidal component. 

Nodes' signals are generated as the mean of the signals on the edges connected to it and, analogously, triangles' signals are generated by taking the mean of the signals of the edges composing it. The task consists of classifying if a signal has or does not have a solenoidal component, i.e. to determine if $A$ in \eqref{eq:sign_synth} is zero or one. We note that nodes and triangles signals are not $\mathbf{s}_{0}$ and $\mathbf{s}_{2}$ from \eqref{eq:sign_synth}; this choice is made to simulate the real scenario in which signals of different simplicial orders come from different sources, and they are not all functions of each other. We split the dataset into training, validation, and test sets, keeping $70\%$ of the data for training, $15\%$ for validation, and the remaining $15\%$ for testing. We train a QSN using the standard cross-entropy loss. A final fully connected readout layer with a softmax non-linearity takes the updated simplices embeddings as input and outputs the class probabilities.  

In Table~\ref{tab:params1}, we report the minimum and the maximum number of parameters for the 2 QSNs while reporting the exact number of parameters per number of layers for the MLP, the GSCN, and the GSAN, per each number of layers. To gain statistical significance, we first generate 10 random datasets (underlying complex plus signals) and we then train the networks using 4 seeds per each of them. 
\begin{table}[!ht]
\centering
\caption{Number of parameters per layers for the architectures applied to Task 1.}
\vspace{6pt}
\label{tab:params1}
\footnotesize
\renewcommand{\arraystretch}{1.2}
\begin{tabular}{c|ccccc}
\toprule
\multirow{2}{*}{Architecture} & \multicolumn{5}{c}{Number of layers} \\
\cmidrule{2-6} 
 & 1 & 2 & 3 & 4 & 5 \\
\midrule
min QSNs & 48 & 83 & 118 & 153 & 188 \\
max QSNs & 70 & 125 & 180 & 235 & 290 \\
MLP   & 69 & 128 & 183 & 243 & 297 \\
GSCN   & 20 & 128 & 180 & 240 & 284 \\
GSAN   & 36 & 128 & 188 & 240 & 292 \\
\bottomrule
\end{tabular}
\end{table}

In Table~\ref{tab:performances1}, we report the average test accuracy, across datasets and seeds,  
against the number of layers for all the models described above. It is evident that QSNs always perform at least as well as classical SNs, outperforming them when the number of layers increases, despite the lower number of parameters. Moreover, quantum SNs show increased stability to depth w.r.t. classical SNs; in particular, while both GSCN and GSAN have a decreasing trend as the number of layers increases, SQSN shows a strongly stable trend, and BQSN an increasing one.
\begin{table}[!ht]
\centering
\caption{Mean test accuracy per layers for the architectures applied to Task 1.}
\vspace{6pt}
\label{tab:performances1}
\footnotesize
\renewcommand{\arraystretch}{1.2}
\begin{tabular}{c|ccccc}
\toprule
\multirow{2}{*}{Architecture}& \multicolumn{5}{c}{Number of layers} \\
\cmidrule{2-6} 
 & 1 & 2 & 3 & 4 & 5 \\
\midrule
SQSN    & 0.949 & $\mathbf{0.956}$ & $\mathbf{0.957}$ & $\mathbf{0.950}$ & $\mathbf{0.951}$ \\
BQSN    & 0.833 & 0.805 & 0.874 & 0.868 & 0.891 \\
MLP     & $\mathbf{0.991}$ & 0.903 & 0.897 & 0.932 & 0.939 \\
GSCN   & 0.939 & 0.872 & 0.798 & 0.736 & 0.598 \\
GSAN    & 0.929 & 0.886 & 0.885 & 0.877 & 0.879 \\
\bottomrule
\end{tabular}
\end{table}

In order to better understand why the SQSN performs better than the BQSN, we performed some further analysis based on the Shannon Entropy\footnote{Here Von Neumann Entropy, which is the quantum counterpart of the Shannon Entropy, would be more appropriate but due to hardware limitations we were not able to compute it reliably.} \cite{LESNE_2014}, useful to quantify the amount of available information in a probability distribution. 

\begin{figure}[!ht]
\centering
\includegraphics[width=.8\textwidth]{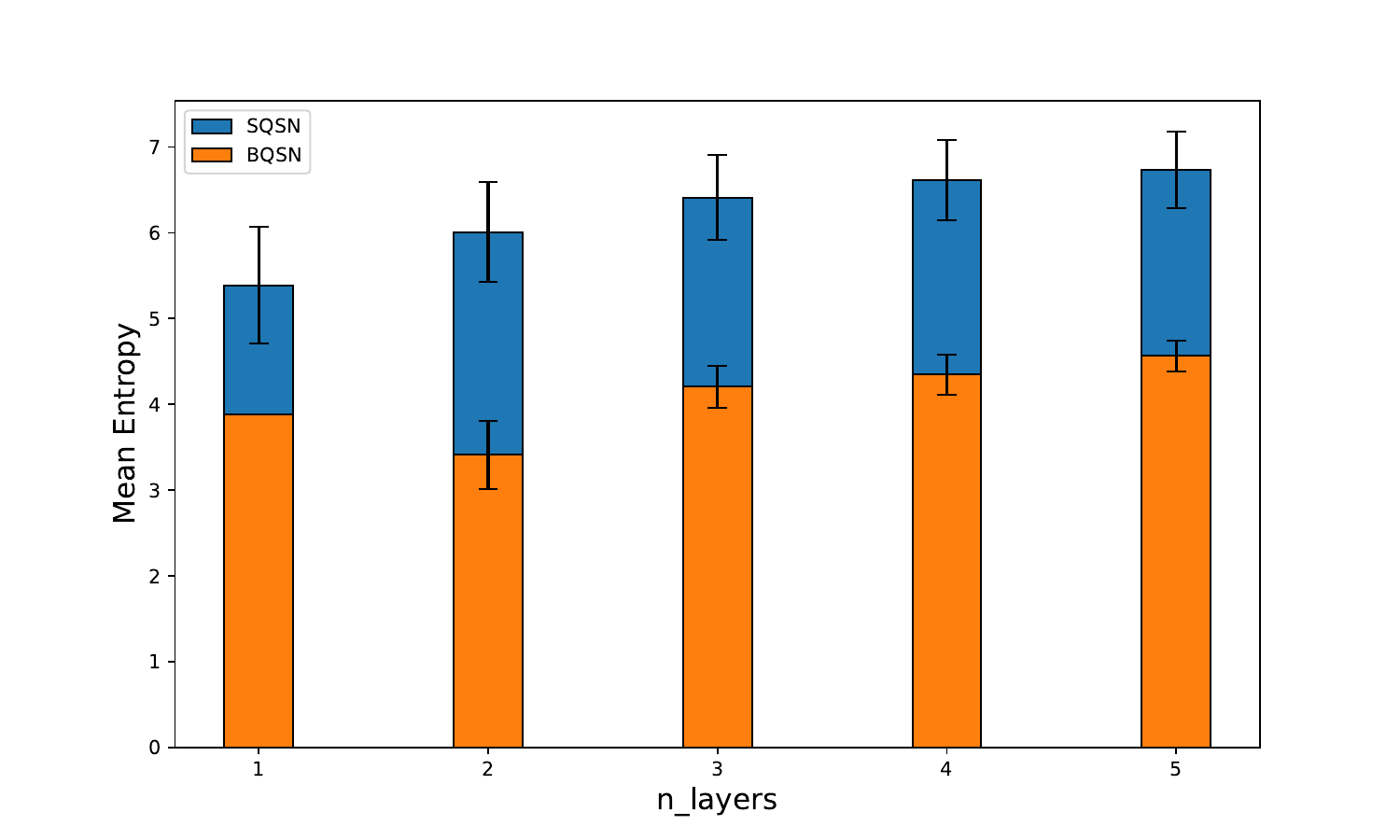}   
\caption{Mean entropy per layers for the quantum architectures applied to Task 1.}
\label{fig:entropy}
\end{figure}

In Fig.~\ref{fig:entropy}, we report the average Shannon Entropy against the number of layers for both BQSNs and SQSNs.
Each QSN has been run $n_\mathrm{shots}=10000$ times for each number of layers with a set of randomly initialized parameters. We emphasize that to make the comparison fair we chose a fixed datapoint in the dataset which initializes the states for each run in the same state. Once the circuit is executed, we compute its output distribution Shannon Entropy, which allows us to obtain the average Shannon Entropy per layer over the $n_\mathrm{shots}$, along with its standard deviation, both reported in Fig.~\ref{fig:entropy}.
As noticeable in the output histogram, the SQSN is able to extract more information on average with respect to the BQSN from the input data, resulting in a better learning performance.

\subsection{Source Localization Problem} 
\label{synthetic_data2}
We consider a synthetic edge classification task. We generate a random undirected graph with $E$ edges using a Stochastic Block Model~\cite{holland1983sbm}. Each graph consists of 6 nodes evenly distributed across 2 communities, with intra-community edge probability set to 0.85 and inter-community edge probability set to 0.25. The intra-community edges are grouped into 2 edge communities, while the inter-community edges form a 3rd group. We create 1000 edge signals, each sampled from a zero-mean Gaussian distribution with a variance of $1/E$. For each signal, we add spikes to randomly chosen source edges within a single community, where the spike intensity follows $\mathcal{N}(0, 1)$. 
The graphs are lifted into simplicial complexes of order two by filling all the triangles, and the spikes are diffused over the graph according to $\mathbf{s}' = \mathbf{S}^t \mathbf{s} + \textbf{n}$, where $\mathbf{S} \in \mathbb{R}^{E \times E}$ is selected as the support of the lower Laplacian of order 1 $\mathbf{L}_1^{(d)}$ (i.e., a binary matrix encoding the lower connectivity of edges simulating a realistic diffusion over a physical/spatial undirected flow network), $t$ is the order of diffusion sampled from a Student-T distribution with 10 degrees of freedom and capped at 100, $\mathbf{s}$ is the original signal with added spikes, and $\textbf{n}$ is additive white Gaussian noise inducing an SNR of 40dB. 

The task is to identify the community responsible for the spikes, making this a classification problem with 3 classes. We split the 1000 signals into training, validation, and test sets, keeping $70\%$ of the data for training, $15\%$ for validation, and the remaining $15\%$ for testing. We train the QSNs using the standard cross-entropy loss. A final fully connected readout layer with a softmax non-linearity takes the updated simplices embeddings as input and outputs the class probabilities.   

In Table~\ref{tab:params2}, we report the number of parameters for the 2 QSNs while reporting the exact number of parameters per number of layers for the classical architectures, per each number of layers. Again, to gain statistical significance, we first generate a random dataset (underlying complex plus signals) and we then train the networks using 6 seeds per each of them.
\begin{table}[!ht]
\centering
\caption{Number of parameters per layers for the architectures applied to Task 2.}
\vspace{6pt}
\label{tab:params2}
\footnotesize
\renewcommand{\arraystretch}{1.2}
\begin{tabular}{c|ccccc}
\toprule
\multirow{2}{*}{Architecture} & \multicolumn{5}{c}{Number of layers} \\
\cmidrule{2-6} 
 & 1 & 2 & 3 & 4 & 5 \\
\midrule
QSNs & 81 & 141 & 201 & 261 & 321 \\
QGNN & 34 & 47 & 60 & 73 & 86\\
MLP   & 69 & 128 & 183 & 243 & 297 \\
GSCN   & 52 & 160 & 212 & 272 & 340 \\
GSAN   & 68 & 160 & 220 & 272 & 324 \\
\bottomrule
\end{tabular}
\end{table}

In Table~\ref{tab:performances2}, we report the average (across datasets and seeds) test accuracy 
against the number of layers for all the models described above. As the reader can notice, Quantum SNs consistently outperform classical SNs, despite the lower number of parameters. 
\textcolor{black}{The results clearly demonstrate that QSNs outperform QGNNs across all configurations of layers. While the QGNN achieves reasonable performances, it consistently lags behind the BQSN in terms of accuracy. In particular, the BQSN shows a significant improvement over the QGNN, with an average enhancement of approximately $9.02\%$ across all layer configurations.
These results underscore the greater effectiveness of QSNs in capturing complex relationships within the data. 
The number of parameters required for 5 layers of QGNN is roughly equivalent to that of 1 layer of QSN. Despite this, we observed that, as the number of layers increases, the performance of QGNNs tends to plateau.
In conclusion, the findings suggest that QSNs, especially the BQSN, are a more effective and efficient choice over QGNNs for this task, achieving better performances even with small graphs, where the influence of higher-order interactions is less relevant.}
\begin{table}[!ht]
\centering
\caption{Mean test accuracy per layers for the architectures applied to Task 2.}
\vspace{6pt}
\label{tab:performances2}
\footnotesize
\renewcommand{\arraystretch}{1.2}
\begin{tabular}{c|ccccc}
\toprule
\multirow{2}{*}{Architecture}& \multicolumn{5}{c}{Number of layers} \\
\cmidrule{2-6} 
 & 1 & 2 & 3 & 4 & 5 \\
\midrule
SQSN    & 0.674 & $\mathbf{0.665}$ & 0.686 & 0.650 & 0.681 \\
BQSN    & $\mathbf{0.681}$ & 0.618 & $\mathbf{0.702}$ & $\mathbf{0.678}$ & $\mathbf{0.701}$ \\
QGNN & 0.608 & 0.578 & 0.659 & 0.593 & 0.665\\
MLP     & 0.394 & 0.531 & 0.630 & 0.583 & 0.606 \\
GSCN   & 0.402 & 0.482 & 0.455 & 0.418 & 0.396 \\
GSAN    & 0.535 & 0.602 & 0.579 & 0.587 & 0.513 \\
\bottomrule
\end{tabular}
\end{table}

\begin{figure}[!ht]
\centering
\includegraphics[width=.8\textwidth]{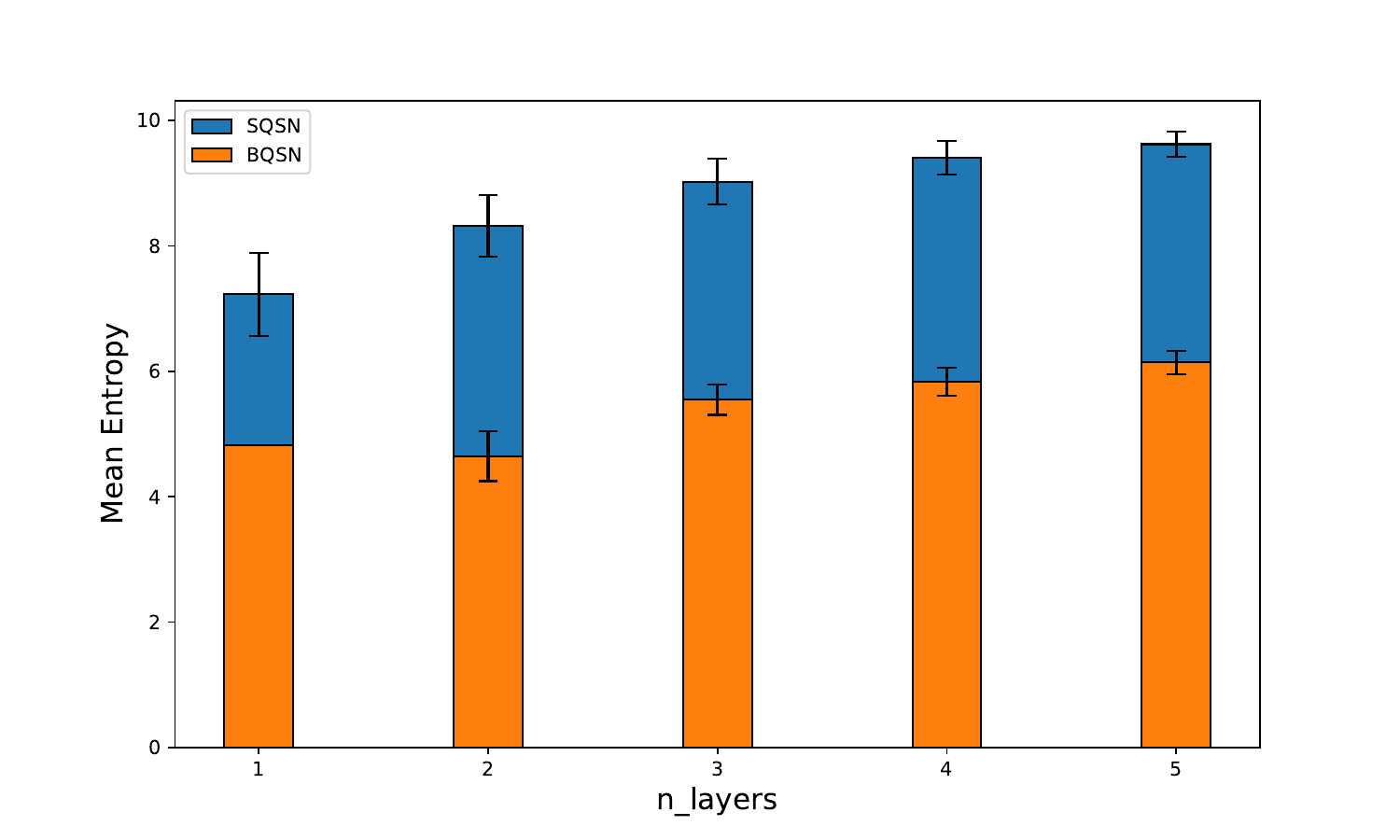}   
\caption{Mean entropy per layers for the quantum architectures applied to Task 2.}
\label{fig:entropy2}
\end{figure}

As for the previous task, in Fig.~\ref{fig:entropy2}  we report the average Shannon Entropy against the number of layers for both BQSNs and SQSNs. The plot is generated with the same procedure.
As expected, also in this case, the SQSN seems able to better extract information, which means the schematic layer is consistent. The observed superior performance of the BQSN over the SQSN likely stems from differences in the nature of the information each model extracts, attributed to variations in their input processing methods. Although the BQSN may capture less comprehensive information overall, the type of information it extracts could be more aligned with the specific task requirements.

\subsection{Experimental Settings}
The models training was conducted using an NVIDIA RTX 3080 Ti with 12 GB of VRAM to ensure efficient simulation while maintaining high computational speed. The QSNs were implemented in Python 3.11.4 with the assistance of Pennylane 0.34.0, and the training process was facilitated using Jax 0.4.17 and Flax 0.7.4, while the classical architectures were implemented and trained with the aid of PyTorch 2.2.2+cu121. The optimizer employed was Adam \cite{adam} with a learning rate of 0.01, incorporating an early stopping mechanism to optimize training efficiency. In all the experiments, the learnable parameters of the QSNs have been randomly initialized from a normal distribution $\mathcal{N}(\mu=0,\, \sigma=\pi)$. Further details regarding the hyperparameters can be found in Table~\ref{tab:hyperparameters}.
\begin{table}[!ht]
\centering
\caption{Hyperparameters used for the for models' training.}
\vspace{6pt}
\label{tab:hyperparameters}
\footnotesize
\renewcommand{\arraystretch}{1.2}
\begin{tabular}{lc}
\toprule
Hyperparameter & Value \\
\midrule
Batch size & 8 \\
Number of epochs & 500 \\
Patience (early stopping) & 50 \\
\bottomrule
\end{tabular}
\end{table}

\section{Conclusion and Future Directions} 


\label{sec:conclusions}
In this work, we introduced Quantum Simplicial Networks (QSNs), QNNs operating on simplicial complexes, being the first Quantum Topological Deep Learning Model. Experiments on synthetic classification tasks showed that QSNs can outperform classical simplicial TDL models in accuracy and efficiency. This paper opens several new venues at the intersection of topological and quantum deep learning. An interesting direction is generalizing QSN to regular cell complexes \cite{sardellitti2022cell}, enabling the quantum modeling of more general higher-order interactions. It would be useful to inject stronger inductive biases in QSNs by forcing some kind of symmetry \cite{mernyei2022equivariant}, e.g. permutation equivariance, or topological property, e.g. Hodge-decomposable embeddings \cite{battiloro2023generalized}. Finally, parameter efficiency could be improved by introducing QSN integrating weight-sharing schemes or having a number of parameters independent of the number of simplices in the complex.




\section*{Acknowledgment}
The contribution of A. Ceschini and M. Panella in this work was in part supported by the ``NATIONAL CENTRE FOR HPC, BIG DATA AND QUANTUM COMPUTING'' (CN1, Spoke 10) within the Italian ``Piano Nazionale di Ripresa e Resilienza (PNRR)'', Mission 4 Component 2 Investment 1.4 funded by the European Union - {NextGenerationEU} - CN00000013 - CUP B83C22002940006.

\bibliographystyle{elsarticle-num}
\setcitestyle{maxbibnames=3, minbibnames=1}
\bibliography{example_paper}
\appendix

\section{Quantum Gates}

Quantum gates are fundamental operations that manipulate the state of qubits within a quantum circuit. These gates can be broadly categorized into single-qubit and two-qubit gates, each serving distinct roles in quantum computation.

\subsection{Single-Qubit Gates}

Single-qubit gates operate on individual qubits and are essential for state preparation, rotation, and superposition. Table~\ref{tab:single_gates} summarizes the most used single-qubit gates.

\begin{table*}[!ht]
 \caption{Single-qubit Gates}

		\renewcommand\arraystretch{1.2} 
		\setlength{\tabcolsep}{6pt}
    \centering
		\resizebox{.8\textwidth}{!}{%
\begin{tabular}{lll}
        \toprule
        Gate & Matrix Representation & Operation \\
				\midrule
        Pauli-X (X-gate) &
        $\begin{bmatrix} 0 & 1 \\ 1 & 0 \end{bmatrix}$ &
        Bit-flip \\[15pt]

        Pauli-Y (Y-gate) &
        $\begin{bmatrix} 0 & -i \\ i & 0 \end{bmatrix}$ &
        Bit-flip and phase-flip \\[15pt]

        Pauli-Z (Z-gate) &
        $\begin{bmatrix} 1 & 0 \\ 0 & -1 \end{bmatrix}$ &
        Phase-flip \\[15pt]

        Hadamard (H-gate) &
        $\frac{1}{\sqrt{2}}\begin{bmatrix} 1 & 1 \\ 1 & -1 \end{bmatrix}$ &
        Superposition \\[15pt]

        X-Rotation $(R_x)$ &
        $\begin{bmatrix} \cos(\theta/2) & -i\sin(\theta/2) \\ -i\sin(\theta/2) & \cos(\theta/2) \end{bmatrix}$ &
        Rotation about X-axis by angle $\theta$ \\[15pt]

        Y-Rotation $(R_y)$ &
        $\begin{bmatrix} \cos(\theta/2) & -\sin(\theta/2) \\ \sin(\theta/2) & \cos(\theta/2) \end{bmatrix}$ &
        Rotation about Y-axis by angle $\theta$ \\[15pt]
				
        Z-Rotation $(R_z)$ &
        $\begin{bmatrix} e^{-i\theta/2} & 0 \\ 0 & e^{i\theta/2} \end{bmatrix}$ &
        Rotation about Z-axis by angle $\theta$ \\[15pt]
        \bottomrule
    \end{tabular}}
		\label{tab:single_gates}
\end{table*}

\subsection{Two-Qubit Gates}
Two-qubit gates manipulate entanglement and interactions between qubits, enabling essential multi-qubit operations in quantum circuits. Some of the key two-qubit gates include CX Gate and SWAP Gate. CX applies a Pauli-X operation to the target qubit when the control qubit is in the $\ket{1}$ state. It serves as a fundamental building block for entanglement generation and conditional operations.
The SWAP gate exchanges the quantum states of two qubits, effectively swapping their values.
The matrix representations and operations for two-qubit gates are presented in Table~\ref{tab:two_gates}.

\begin{table*}[!ht]
    \caption{Two-qubit gates}
		\renewcommand\arraystretch{1.2} 
		\setlength{\tabcolsep}{6pt}
    \centering
		\resizebox{.8\textwidth}{!}{%
    \begin{tabular}{lll}
        \toprule
        Gate & Matrix Representation & Operation \\
				\midrule
        CNOT (Controlled-X gate) &
        $\begin{bmatrix} 1 & 0 & 0 & 0 \\ 0 & 1 & 0 & 0 \\ 0 & 0 & 0 & 1 \\ 0 & 0 & 1 & 0 \end{bmatrix}$ &
        Entangles if control qubit is $|1\rangle$ \\[15pt]
				
        SWAP &
        $\begin{bmatrix} 1 & 0 & 0 & 0 \\ 0 & 0 & 1 & 0 \\ 0 & 1 & 0 & 0 \\ 0 & 0 & 0 & 1 \end{bmatrix}$ &
        Swaps states of two qubits \\[15pt]
        \bottomrule
    \end{tabular}}
		\label{tab:two_gates}
\end{table*}



\end{document}